\documentclass[conference]{IEEEtran}
\usepackage{algorithmic}
\usepackage{algorithm}           
\usepackage{graphicx}
\usepackage{amsmath}
\usepackage{array}
\usepackage{multirow}
\usepackage{float}

\begin{document}
%
\title{A coarse-to-fine algorithm for registration in 3D street-view cross-source point clouds}


\author{Anonymous}
\author{\IEEEauthorblockN{Xiaoshui Huang$^1$, Jian Zhang$^1$,  Qiang Wu$^1$, Lixin Fan$^2$ and Chun Yuan$^3$}
\IEEEauthorblockA{$^1$Global Big Data Technologies Centre, School of Computing and Communication.\\
University of Technology Sydney,  Australia}
\IEEEauthorblockA{Emails:\{Xiaoshui.Huang@student., Jian.Zhang@, Qiang.Wu@\}uts.edu.au}

\IEEEauthorblockA{$^2$Nokia Technologies. Tampere, Finland \\
	Email: Lixin.Fan@nokia.com}

\IEEEauthorblockA{$^3$Graduate School at Shenzhen, Tsinghua University, Shenzhen, China\\
	Email: yuanc@sz.tsinghua.edu.cn}
}

\maketitle

\begin{abstract}
	
With the development of numerous 3D sensing technologies, object registration on cross-source point cloud has aroused researchers' interests. When the point clouds are captured from different kinds of sensors, there are large and different kinds of variations. In this study, we address an even more challenging case in which the differently-source point clouds are acquired from a real street view. One is produced directly by the LiDAR system and the other is generated by using VSFM software on image sequence captured from RGB cameras. When it confronts to large scale point clouds, previous methods mostly focus on point-to-point level registration, and the methods have many limitations.The reason is that the least mean error strategy shows poor ability in registering large variable cross-source point clouds. In this paper, different from previous ICP-based methods, and from a statistic view, we propose a effective coarse-to-fine algorithm to detect and register a small scale SFM point cloud in a large scale Lidar point cloud. Seen from the experimental results, the model can successfully run on LiDAR and SFM point clouds, hence it can make a contribution to many applications, such as robotics and smart city development.

\end{abstract}

\IEEEpeerreviewmaketitle

\section{Introduction}
Researchers have shown great interests in the applications of registration, such as view searching in smart city, location-based service, street view reconstruction and augmented reality. With the progress of sensing technology, many types of 3D point cloud sensor have been developed. Coming from different types of sensors, point clouds are cross-source ones. Compared to same source, registration on cross-source point clouds show great generalization. In this paper, we propose a method to conduct the registration of a small-scale SFM point cloud on a large-scale street-view point cloud.

\begin{figure}[t]
	\centering
	\includegraphics[width=8.5cm,height=6.0cm]{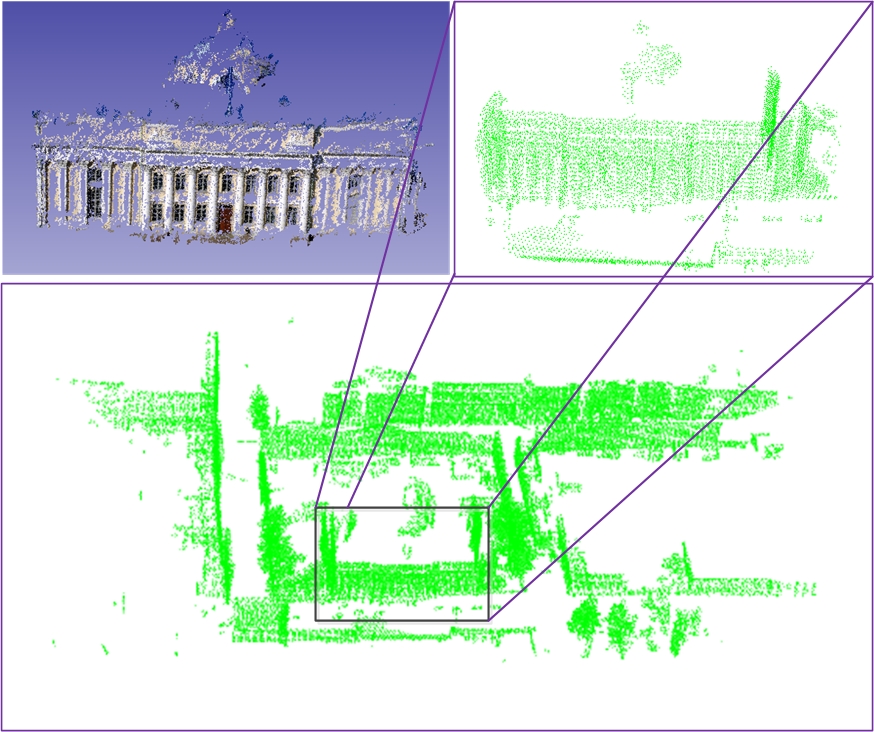}
	\caption{ An example of cross-source point clouds of SFM and LiDAR highlighted from the street view scene. The top left is the SFM point cloud and the top right is the detected registration result on LiDAR point cloud.}
	\label{f1}
\end{figure}

Figure \ref{f1} is a typical example of cross-source point clouds, which contains cross-source problems. At least four challenges can be posed in solving cross-source point cloud registration. (1) Density variation. Different sampling density and sampling theory from various sensors result in distinguished point number in two types of point clouds. Therefore, one point cloud may be much denser than the others. (2) Scale variation. Due to their different sampling density, the scale is hard to maintain same metric in two types of sensor. Furthermore, in terms of the point clouds reconstructed by structure from motion software (e.g. VSFM\cite{vsfm}) or SLAM \cite{huang2016real}, the scale information is usually unknown, thus we need to estimate the scale. (3) Noise, outliers and missing data. Different sensing mechanisms make a large amount of noise and outliers in the cross-source point clouds; and some parts of the scenes cannot be produced points in point clouds. For example, VSFM is unable to generate points in uniform image. (4) Viewpoint variation. Viewpoint divergence, which is normal in cross-source point clouds, makes the describing content originally different to some extent. 

With respect to point cloud registration, the existing methods can be casted into two categories: direct and transformation methods. Among them, direct methods directly use the coordinate of point clouds and extract some descriptors to assistant matching and registration, with typical examples of ICP \cite{icp} and other feature-based methods \cite{hontani2010point,sharp2002icp,huang2015dense,huang2015graph} . These methods show great ability in same source point clouds while different extent of limitation on cross-source point clouds. To be specific, they are aimed at looking for an exact matching point for each point, which takes up only a very small proportion in cross-soured point clouds though. Consequently, direct methods make matcher point searching confused all the time when there is a large amount of outliers and noises. In addition, they rely on initialization. For example, similar to our work, Peng. et al \cite{7025406} proposed two-stage algorithm by using ICP. Due to the ICP's limitation, the final registration results accuracy is low, which is visually not registered correctly.  In terms of transformation methods, they firstly transform 3D point clouds into other space or other model, and then use these transformed data for matching and registration, with typical example of Gaussian mixture model (GMM) \cite{JR-MPC,jian2011robust}, using GMM to describe point cloud and matching them. These methods focus more on global information while ignore local structure distortion. It shows many advantages than direct methods in dealing with cross-source problems. Our algorithm belongs to transformation methods.

In this paper, a novel coarse-to-fine algorithm is proposed to register two cross-source point clouds (one is whole street, the other is a small part) . There are mainly two steps: 1) top K potential regions are detected by a coarse matching in the large-scale LiDAR street-view point cloud for a small-scale SFM point cloud. 2) a generative GMM registration method is applied to refine the matching results from the first stage.

The main contribution is a effective coarse-to-fine pipeline which utilizes the concept of GMM to do registration in large-scale cross-source point clouds. Different to previous generative GMM, we propose a coarse-to-fine, scale normalization and complexity reduction strategies to extend it to be suitable to the new cross-source problem. Different to previous ICP-based method \cite{7025406}, we utilize the statistic property of cross-source point cloud to deal with the large variations in local points. It successfully overcomes the point-to-point limitation in ICP-based methods.

\section{Coarse-to-fine Algorithm}
\begin{figure}[h]
	\centering
	\includegraphics[width=8.5cm,height=3.0cm]{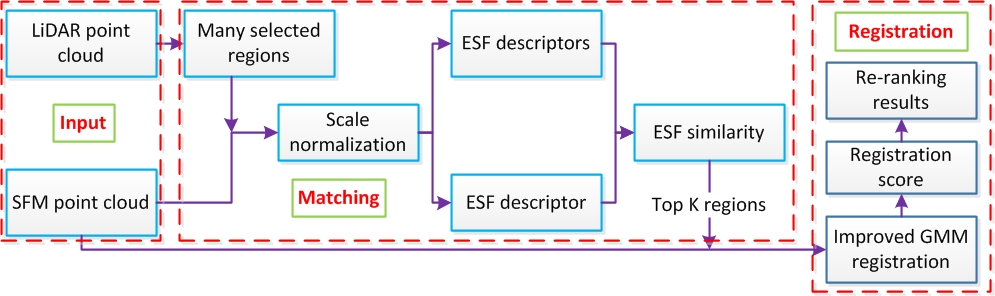}
	\caption{Overview of the proposed algorithm.}
	\label{f2}
\end{figure}

The algorithm is illustrated in Figure \ref{f2}. It contains coarse matching and fine registration, among which coarse matching aims at finding the top $K$ potential regions in LiDAR point cloud that potentially match with SFM point cloud. It substantially reduces the number of candidate regions and hence saves computation cost of the next stage. We compute ESF ( Ensemble of Shape Functions) descriptors \cite{ESF} of these potential regions and use them to conduct the first coarse matching. Then, a improved generative GMM-based registration is performed to obtain the transformation of two cross-source point clouds and use the transformation error to refine the matching results. Furthermore, main steps of the second stage are (1) obtain transformation matrix of each registration; (2) acquire residual error of each registration by applying transformation matrix to the original two cross-source point clouds (e.g. selected LiDAR region and SFM); (3) use residual error to re-rank the matching results and output the ranked registration results. After registration, accurate transformation matrix is obtained, and they can be used for applications such as location based service.

\subsection{Coarse Matching}
This part is aimed at computing Top K potential regions from LiDAR point cloud for the SFM point cloud. We compute ESF descriptor for point clouds regions (LiDAR and SFM point clouds' region). ESF is a statistical descriptor depicting a point cloud globally. As a 640-bin histogram, it describes the point cloud based on ten different kinds of shape distribution, including three kinds of line distance distribution, three kinds of angle distribution, three kinds of triangle area distribution and one kind of ratio of line distance distribution \cite{ESF}. Each shape distribution is sampled by a 64-bin histogram and all the ten histograms are concatenated to form the ESF descriptor. The ESF is robust to local structure distortion, which is a valuable advantage in describing high noisy cross-source point clouds (street view point clouds are employed in our experiments).

Since scale difference is a common variation in cross-source point clouds, we should remove it before ESF computation. As to matching and registration, our goal is to find a region with largest similarity, thus the matched point clouds should have a large proportion of overlapping. In this paper, we assume that the matched point clouds have the same voxel box, and then the scale can be computed automatically by

\begin{eqnarray}
scale=radius\__{lidar}/radius\__{SFM}
\end{eqnarray}

where $radius\__{lidar}$ is the radius of voxel box containing LiDAR point cloud while $radius\__{SFM}$ is the radius of voxel box containing SFM point cloud. For $radius\__{SFM}$, it can be computed from the point cloud by counting the maximized distance to  the center point. For $radius\__{lidar}$,  in order to cover the potential regions of SFM point clouds, multi-scale radius is needed (e.g. the LiDAR radius is defined from 10 meter to 1 kilometer). A case of multi-scale radius selection is shown in our experiments.

With the scale factor computed, a scale transformation to all points in SFM point cloud is carried out.

\begin{eqnarray}
P_{SFM}=P_{SFM}*scale
\end{eqnarray}

where $P_{SFM}$ represents the SFM point cloud. After scale normalization, ESF is applied to both LiDAR and SFM point clouds. Then we compute the similarity($Simi\__{matching}$) between the ESF descriptors by

\begin{equation}
Simi\__{matching}=\Arrowvert ESF\__{lidar} - ESF\__{SFM}\Arrowvert_F
\end{equation}

where $ESF\__{lidar}$ is the ESF descriptor of selected region in LiDAR point cloud, $ESF\__{SFM}$ is the ESF descriptor of SFM point cloud. For SFM point cloud, top $K$ potential regions in LiDAR point cloud are selected by using these similarity values.  In this stage, only top $K$ potential regions are selected, however, the accurate registration relations (transformation matrix) is still unknown.

\subsection{Fine Registration}
To obtain the correct registration results for SFM point cloud from detected top $K$ regions in the large LiDAR point cloud, the fine registration is indispensable. For cross-source point cloud registration, the conventional point-to-point level methods face much difficulty in registering these large variable cross-source point clouds. This is because of their simple least square mean error of point-level correspondence can be easily lead to sub-optimal in the large variant cross-source point clouds. In order to address the limitation issues of previous direct methods in terms of cross-source problems, we use GMM to consider global statistical properties (e.g. global shape or distribution). GMM-based method focuses on whether the two cross-source point clouds are globally registered and ignores the large variation in local structure.

A generative GMM method is presented by considering multiple point clouds from a same GMM named JR-MPC \cite{JR-MPC}. The paper only reports experiments on same source. As the two registered  cross-source  point clouds  are depicting for the same region (e.g. a house or a tree), it is also reasonable to use a GMM to describe the same region and consider these point clouds as two samples for this GMM. Hence, in this paper, we extend this method to cross-source point cloud registration problem. In the following part, we will describe the proposed strategies to extend JR-MPC to cross-source registration problem.

Due to the cross-source problems (discussed in section 1), the above JR-MPC cannot apply to cross-source point cloud registration problems directly. There are two main problems: 1) JR-MPC are originally designed to address multi point sets registration problem. It has the assumption that the point sets under the same scale. 2) In the expectation step of these generative GMM methods, it need to estimate probability of each point belonging to every Gaussian model. The computation and memory complexity are very large which is $O(M*K+N*K)$, where $M$ and $N$ are the number of two point sets and $K$ is the Gaussian model. In the model of JR-MPC, even worse, the complexity is  $O(M*N*K)$. If considering $M\approx N$,  the complexity both in memory and computation is approximately $O(N^2*K)$, which is prohibitive for large scale cross-source point cloud. We will describe how to effectively deal with these problems.

Firstly, for the scale problem, it have been normalized in the previous coarse matching stage. Due to the proposed coarse matching stage, it not only provides the top $K$ potential regions in the large scene, but also normalize the scale problem that suitable for the registration stage.

\begin{figure}[h]
	\centering
	\includegraphics[width=8.5cm,height=3.0cm]{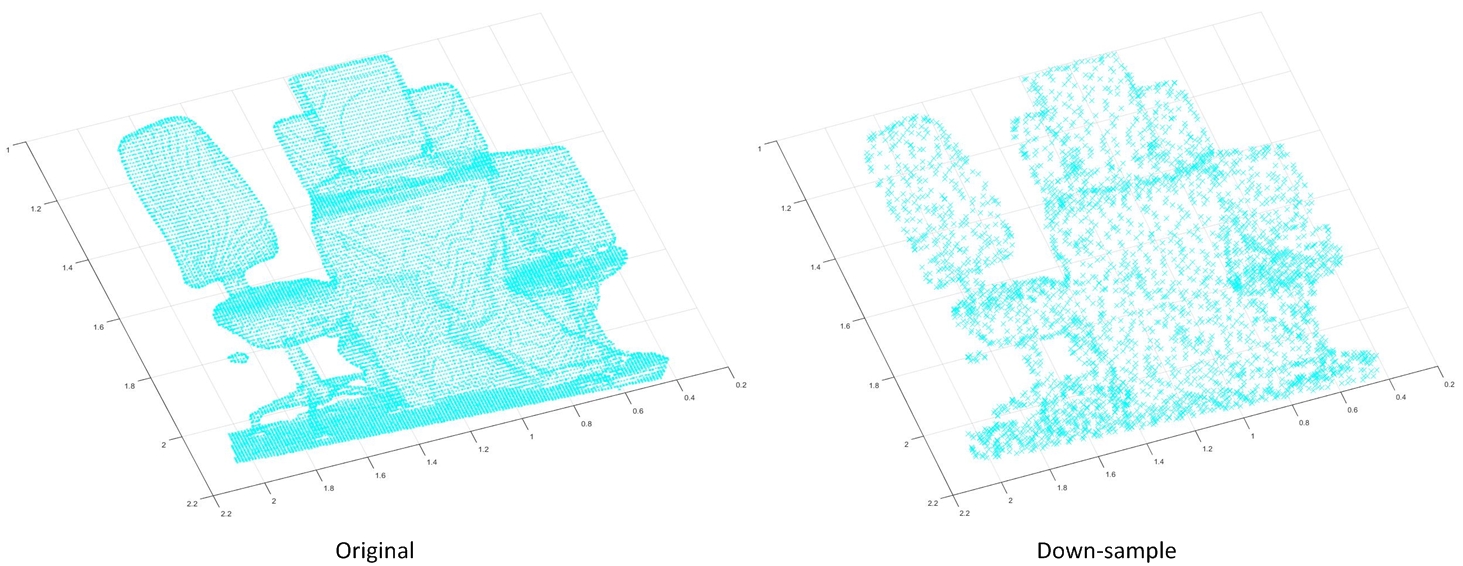}
	\caption{Visual results of original and down-sample point clouds.}
	\label{downsample}
\end{figure}

Secondly, for the complexity problem, we need to look deeply to the cross-source problem so as to settle the computation and memory complexity problem. Due to large variations in cross-source point clouds, the exact matched points are only a small proportion. So, trying to find exact matched point is a difficult task in changeable data. Instead of using the original changeable points, the statistical property (GMM) shows stable and more value in registering two cross-source point clouds. It avoids the local variations and focus on more on the global mean and variance. Also, in this paper, the cross-source point cloud registration problem only contains rigid transformation. Hence, we uniformly down-sample the point cloud when the points are over two thousand. After down-sampling, the global shape or structure and rigid transformation still keep the same as the original point cloud (see Figure \ref{downsample}). Due to the shape of two point clouds are all kept, the region of GMM depicted are the same. So, the transformation matrix computed by the down-sample point cloud is the same to the original point cloud. In this way, these two point clouds registration can be successfully converted from a large complexity problem to a feasible problem. If a rigid transformation is computed by using down-sampled point cloud, we can directly apply the rigid transformation to the original point clouds and obtain the final registration results. 

In the fine registration stage, when the improved JR-MPC is completed, a residual error is computed to re-rank the matching results in the first stage. To compute the residual error, the computed transformation matrix from revised JR-MPC is applied to perform transformation to the original point clouds. Next, the nearest neighbour is computed for each point and the mean residual is computed following. The residual error is computed by

\begin{eqnarray}
R\__{score}=\sum\limits_{i}^{N}\Arrowvert m_i-T(d_i)\Arrowvert_2
\end{eqnarray}
where $m_i$ is the $i^{th}$ point in point cloud $A$; $d_i$ is the nearest neighbor of $m_i$ in the matched point cloud $B$. A lower $R\__{score}$ means the two point clouds are more similar.However, based on our observation, the $R\__{score}$ always shows lower value in small-scale point clouds. To eliminate this scale bias, a penalty is defined related to the scale value:

\begin{eqnarray}
Final\__{score}=exp(-\frac{scale^2}{\alpha})*R_{score}
\end{eqnarray}
where, $exp(-\frac{scale^2}{a})$ is the penalty for scale variation. $\alpha$ is the parameter to control the penalty, scale is computed by formulation (1). The final ranking regions are sorted by the $Final\__{score}$ value and the top ranked one represents the best matching to the SFM point cloud. The pseudo-code of the coarse-to-fine algorithm is shown in Algorithm \ref{alg:TS}. 

\begin{algorithm}[htb]         
	\caption{ Pseudocode of coarse-to-fine algorithm }             
	\label{alg:TS}                  
\begin{algorithmic}
		
	\REQUIRE  $cross\ sourced \ point\ clouds$
	\ENSURE $Top\ 5\ Registered\ regions$
	\STATE $\textbf{Matching}:$
	\STATE $\ \ \ \ 1.\ Select\ multi-scale\ regions\ from\ LiDAR$
	\STATE $\ \ \ \ 2.\ Scale\ normalization\ by\ Eq. (2)$
	\STATE $\ \ \ \ 3.\ Compute\ ESF\ for\ these\ regions$
	\STATE $\ \ \ \ 4.\ Select\ Top\ K\ regions\ by\ Eq. (3)$
	\STATE $\textbf{Registration}:$
	\STATE $\ \ \ \ 5.\  Down - sample\ point\ cloud$
	\STATE $\ \ \ \ 6.\  Compute\ Transformation\ T\ by\ JR-MPC$
	\STATE $\ \ \ \ 7.\ Compute\ Final\__{score}\ by\ Eq. (7) $
	\STATE $\ \ \ \ 8.\ Re-ranking\  using\ Final\__{score}$
	\STATE $\ \ \ \ 9.\ Cut\ off\ at\ Top\ 5$
\end{algorithmic}
\end{algorithm}

\section{Experimental results}

The experiments are conducted on real cross-source point clouds that are combined by LiDAR and SFM point cloud. LiDAR point clouds are captured from three different scenes in Helsinki (Helsinki Cathedral, Helsinki station and Library of University of Helsinki), with hundreds of millions of points on each original LiDAR point cloud. To efficiently match and register on the large volume data, the LiDAR point clouds are down-sampled into 10\% of the original points. For SFM point clouds, three typical buildings are selected and 2D images are captured by digit camera. Helsinki Station is divided into two objects: station south and station east.

\begin{figure}[h]
	\centering
	\includegraphics[width=8.5cm,height=8.8cm]{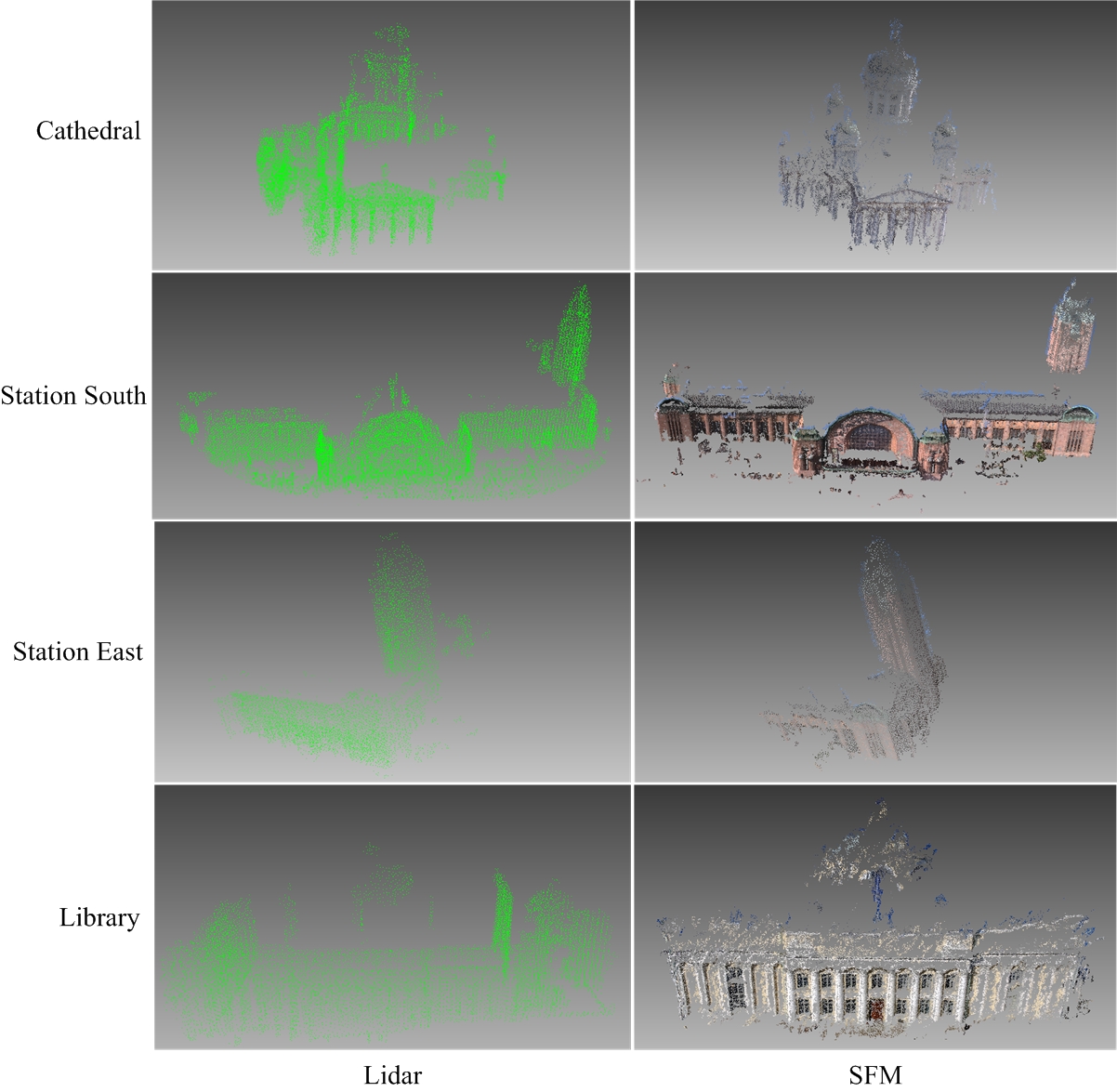}
	\caption{Eight point clouds of four objects named Cathedral, Station south, Station east and Library. Each row represents one object with two cross-source point clouds. The left column is LiDAR and the right is SFM.}
	\label{sourcePC}
\end{figure}

We use 2D images and VSFM \cite{vsfm} to build a software-reconstructed point clouds. The four objects of LiDAR and SFM point clouds are illustrated in Figure \ref{sourcePC}. Before applying the proposed algorithm, standard pre-processing, such as removal of sparse outliers, is conducted for both point clouds. Considering computation complexity reduction, the performance of the proposed algorithm is evaluated on a subset data. The subset data is generated by 7 different scale spheres scanning all the LiDAR point clouds. The radius of the spheres ranges from 30 to 60 with an interval of 5. A hundred regions are selected under each scale. The subset data are regarded as candidate regions for matching and registration. The candidate regions will cover more than $50\%$ areas of LiDAR point clouds. The matching and registration is then regarded as a retrieval problem.The target point cloud(SFM) is retrieved from the 700 candidate LiDAR regions (100 candidates for each one of the 7 scales).

Based on our study, the first matching stage can achieve the best performance when the number of ESF sampling level is 64. The number of potential regions kept for the second fine registration stage $K$ is selected as 20 in all the experiments.

We define two single stage baseline systems and select ESF+ICP, ESF+GO-ICP as our compared methods. For baseline systems, like \cite{7025406}, one is retrieved by ESF only to measure the ESF similarity of point clouds. The other is applying ICP to compute the residual error on each region in every point clouds. Since scale variation is a common problem existing in cross-source point cloud, we normalize the scale by our scale estimation method before applying ICP. In the baseline system, one difference is that the scales do not adjust the residual produced by ICP, as that in the proposed method. To compare the performance of proposed method, we regard the candidate regions which cover $>90\%$ area of the target object and $<10\%$ points associated with the background are regarded as ground-truth data. In this paper, rank-5 measurement is proposed and the ground-truth number is more than 5. According to their ESF similarity or final residual error, candidate regions selected from LiDAR point clouds are sorted and the rank is cut off at top 5. The algorithm shows better performance when there are more retrieved ground-truth regions. All experiments are conducted in a computer with 4-core 3.2GHz CPU and 8GB memory. The results are illustrated in Table \ref{t1}.

\begin{figure*}[t]
	\centering
	\includegraphics[width=18cm,height=10.0cm]{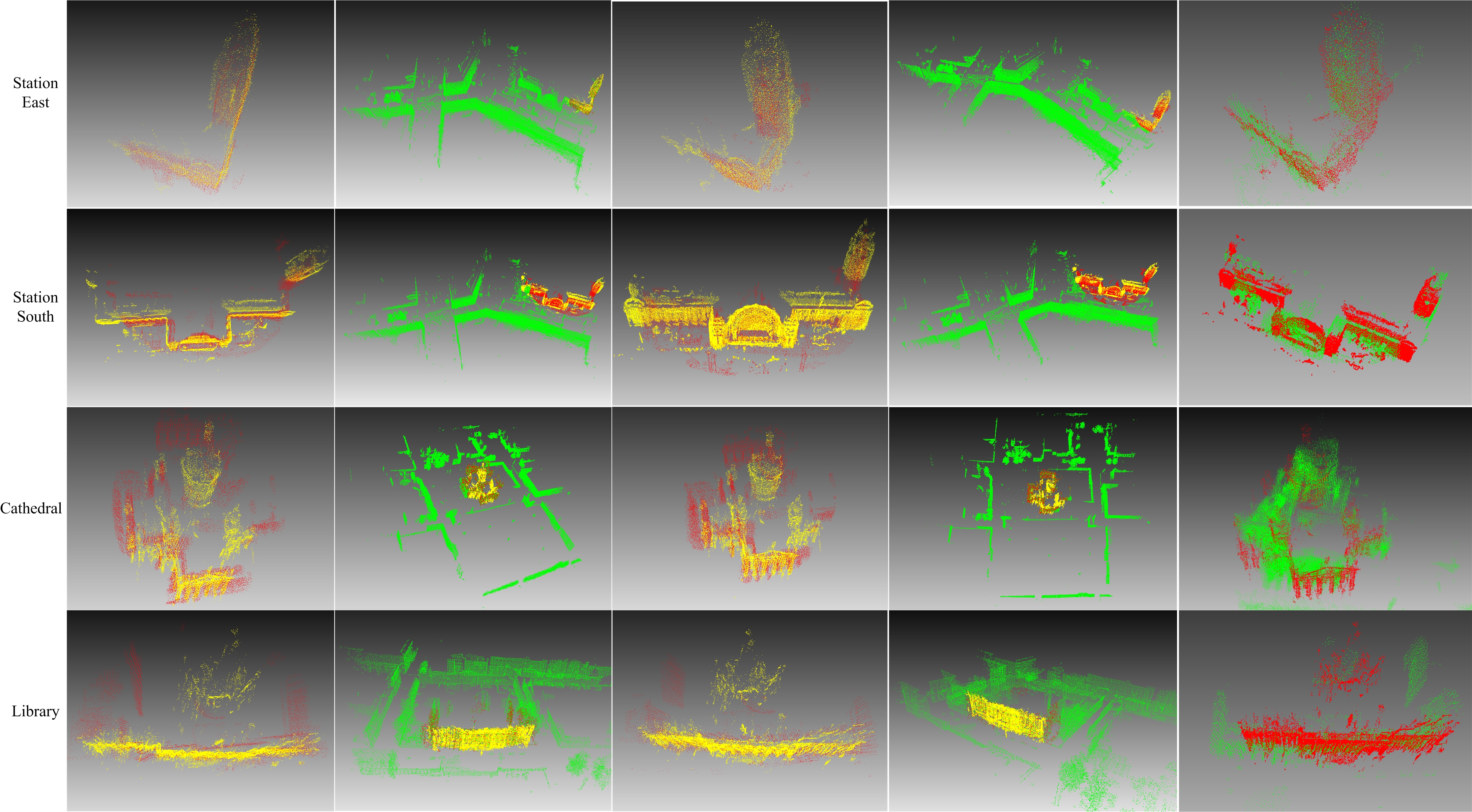}
	\caption{The top 2 registration results of the 4 objects with the proposed method and \cite{7025406}. Each row represents the results for one object. Figures in the 1th column and the 3th respectively represent the GMM registration results of retrieved rank 1 and rank 2 candidate regions in LiDAR and SFM point cloud. Figures in the 2th to 4th columns respectively represent the registration results of whole LiDAR street scene and SFM point cloud. The target in LiDAR is in red, the reference in yellow and other points are in green. Figure in the 5th column is the top 1 registration result of \cite{7025406}, with LiDAR in green and SFM in red. } 
	\label{results}
\end{figure*}
\begin{table*}[ht]
	\centering
	\caption{The performance of the proposed method and the compared methods} 
	\begin{tabular}{|p{4cm}|p{1.2cm}|p{1.2cm}|p{1.2cm}|p{1.2cm}|p{1.2cm}|p{1.2cm}|p{1.2cm}|p{1.2cm}|}
		\hline 
		\multirow{2}{*}{} & \multicolumn{2}{c|}{cathedral} &  \multicolumn{2}{c|}{library} & \multicolumn{2}{c|}{station south} &  \multicolumn{2}{c|}{station east} \\
		\cline{2-9} 
		& accuracy & time(s) & accuracy & time(s) & accuracy & time(s) & accuracy & time(s) \\
		
		\hline 
		Baseline: single stage ESF & 4 & 24 & 0 & 25 & 2 & 23 & 0 & 25 \\
		\hline
		Baseline: single stage ICP & 5 & 305 & 5 & 241 & 0 & 167 & 5 & 522 \\
		\hline
		ESF-64 + ICP without adjusting final residual \cite{7025406} & 5 & 85 & 3 & 73 & 0 & 56 & 4 & 139 \\
		\hline
		ESF-64 + ICP with adjusting final residual \cite{7025406} & 5 & 85 & 3 & 73 & 4 & 56 & 5 & 129 \\
		\hline
		ESF-64 + Go-ICP with adjusting final residual \cite{7025406} & 5 & 85 & 3 & 73 & 4 & 56 & 5 & 129 \\
		\hline
		The proposed method & 5 & 300 & \textbf{5} & 223 & \textbf{5} & 320 & 5 & 256 \\
		\hline
		
	\end{tabular} 
	\label{t1}  
\end{table*}

As shown in Table \ref{t1}, the single stage ESF performs faster but suffers from low accuracy. The baseline of single stage ICP, however, possesses higher accuracy, but it is the most time consuming method. The compared method \cite{7025406} runs much faster than the baseline of single stage of ICP and the proposed method. It uses ESF to quickly remove many incorrect candidates, and then ICP is applied to refine the result, saving a large amount of time. However, it shows lower accuracy to the proposed method, which can be visually seen from Station South and Cathedral in Figure \ref{results}. Using ESF as well, the generative GMM in the proposed method is based on the assumption that two point clouds come from the same object. If the two point clouds have plenty of differences, they are original not registered and it shows high residual error. Also, library and station south results in Table \ref{t1} show that the proposed algorithm is more robust in registration. The proposed method can be robust in detecting the top 5 ground-truth regions from cross-source point clouds (as described before, the ground-truth regions are more than 5). Therefore, the proposed method not only retrieves the correct regions but also registers them more accurately than the compared methods.

In addition, the proposed method is conducted on the whole data set, where multi-scale regions over the whole scene are tested, which means that much more negative regions are included. The top two registered point clouds of each object are presented in Figure \ref{results}. It shows the high registration accuracy of the proposed method from the viewpoint, especially on Station South and Cathedral.

\section{Conclusion}

In this paper, a novel coarse to fine algorithm is proposed to address the problem of cross-source point cloud registration. In the fist stage, coarse matching is performed to quickly detect a few potential matched regions. In the second stage, two revisions about scale and complexity are proposed to extend the recent generative Gaussian mixture model method to cross-source point cloud registration problem. It refines the matching results and accurately finds out registration regions. The proposed method does not rely on least square mean error, but rather utilizes the statistical property that is robust to the cross-source problems. It can efficiently detect the potential regions from a large scene and register them accurately. The proposed method can not only detect where the reference point cloud is located in the big scene but also obtain the accurate pose related to the big scene. The future work is to develop many applications with this method in areas such as location-based service in smart city and robotics.

\bibliographystyle{IEEEtran}
\bibliography{IEEEfull}

\end{document}